\documentclass[letterpaper, 10 pt, conference]{ieeeconf}  

\IEEEoverridecommandlockouts                             
\usepackage{graphics}
\usepackage{siunitx}
\usepackage{epsfig} 
\usepackage{mathptmx} 
\usepackage{times} 
\usepackage{amsmath} 
\usepackage{amssymb}  
\usepackage{subcaption}
\usepackage{graphicx}
\usepackage{float}
\usepackage{caption}
\usepackage{mathtools}
\usepackage{stfloats}
\usepackage{mathrsfs}
\usepackage{multirow}
\usepackage{makecell}
\usepackage{vcell}
\usepackage{booktabs}
\usepackage{diagbox}
\usepackage{csquotes}
\usepackage{cite}

\usepackage[hyphens]{url}
\usepackage{cite}

\usepackage{hyperref}
\hypersetup{
colorlinks=true,
linkcolor=blue,
filecolor=magenta,
urlcolor=black,
}

\title{\LARGE \bf
External Human-Machine Interface on Delivery Robots: Expression of Navigation Intent of the Robot
}
\author{Shyam Sundar Kannan$^{1}$, Ahreum Lee$^{2}$, and Byung-Cheol Min$^{1}$
\thanks{$^{1}$Shyam Sundar Kannan and Byung-Cheol Min are with SMART Lab, Department of Computer and Information Technology, Purdue University, West Lafayette, IN 47907, USA \tt\small{kannan9@purdue.edu | minb@purdue.edu}}%
\thanks{$^{2}$Ahreum Lee is with the University of Eastern Finland, Joensuu, Finland {\tt\small ahreum.lee@uef.fi}}%
}
\begin{document}
\maketitle
\thispagestyle{empty}
\pagestyle{empty}

\begin{abstract}
External Human-Machine Interfaces (eHMI) are widely used on robots and autonomous vehicles to convey the machine's intent to humans. Delivery robots are getting common, and they share the sidewalk along with the pedestrians. Current research has explored the design of eHMI and its effectiveness for social robots and autonomous vehicles, but the use of eHMIs on delivery robots still remains unexplored. There is a knowledge gap on the effective use of eHMIs on delivery robots for indicating the robot's navigational intent to the pedestrians. An online survey with 152 participants was conducted to investigate the comprehensibility of the display and light-based eHMIs that convey the delivery robot's navigational intent under common navigation scenarios. Results show that display is preferred over lights in conveying the intent. The preferred type of content to be displayed varies according to the scenarios. Additionally, light is preferred as an auxiliary eHMI to present redundant information. The findings of this study can contribute to the development of future designs of eHMI on delivery robots. 
\end{abstract}

\section{Introduction}
Recent years have seen the emergence of delivery robots in our day-to-day environment. Delivery robots are being used for the delivery of food and packages. These robots navigate on the sidewalks along with the pedestrians. The robots and humans navigate within close proximity and require safe and efficient navigation. Humans use bidirectional non-verbal communication to express their directional intent while walking on the sidewalk \cite{wood2015interpersonal}. Humans express their navigation intent through various cues such as gestures, gazes, and facial expressions \cite{birdwhistell2010kinesics}. This non-verbal communication enables the smooth navigation of humans. Delivery robots are limited in their ability to make or perceive these non-verbal communications. This limits the delivery robots from making smooth navigation amongst pedestrians which is an essential part of social navigation. 

\begin{figure}[t]
    \centering
    \includegraphics[width=0.35\textwidth]{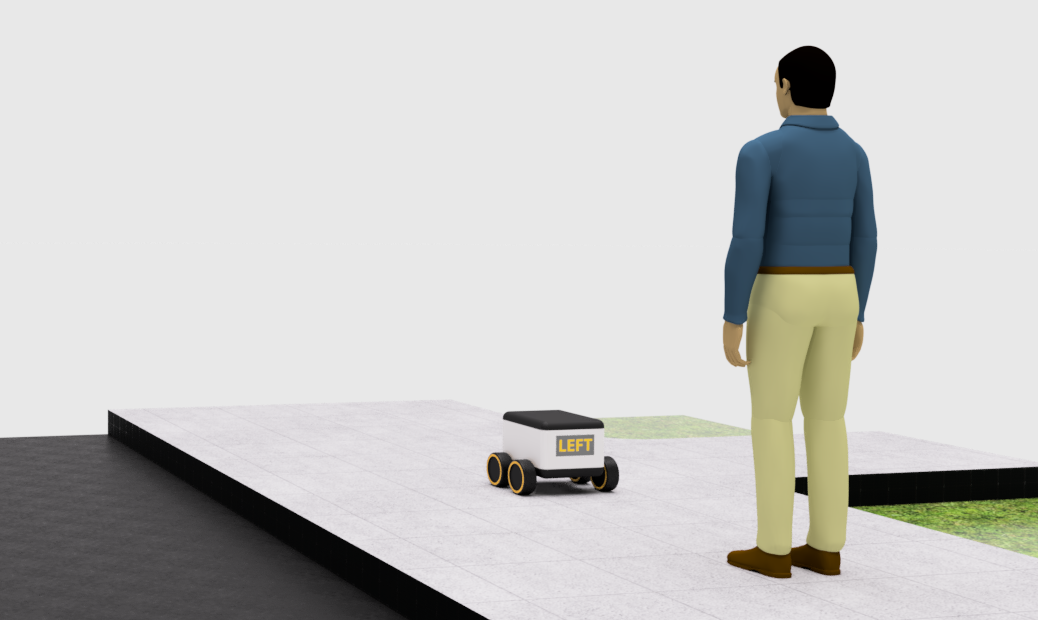}
    \caption{A Delivery robot interacting with a pedestrian in close proximity and using a text on the display to signal its turning intent. }
    \label{fig:intro_figure}
    \vspace{-7mm}
\end{figure}

Current researches have focused on the development of external Human-Machine Interfaces (eHMIs) for the robots to communicate their intent to the humans \cite{matsumaru2006mobile, shrestha2018communicating, fernandez2018passive, tafesse2018analysis}. Various eHMI modalities like light, display, and projection on the floor have been explored. The existing works have mainly focused on an indoor setting for robots like autonomous wheelchair \cite{watanabe2015communicating}, indoor service robots \cite{baraka2016enhancing}, and so on. However, there is no consensus regarding the eHMI design for delivery robots that primarily operate outdoors. The findings of the current research may not be well translated to a delivery robot due to inherent differences in the operational environment. The delivery robots should be able to handle various dynamics of the outdoor world like crowd \cite{trautman2010unfreezing} as well as vehicle traffic for the fast and safe delivery of goods. Hence, it is important to investigate and develop eHMIs that are specifically designed for delivery robots. This is critical to enable smooth navigation of both the pedestrian and the delivery robot.

\begin{figure*}[t]
\begin{subfigure}{.315\textwidth}
  \centering
  \includegraphics[width=.92\linewidth]{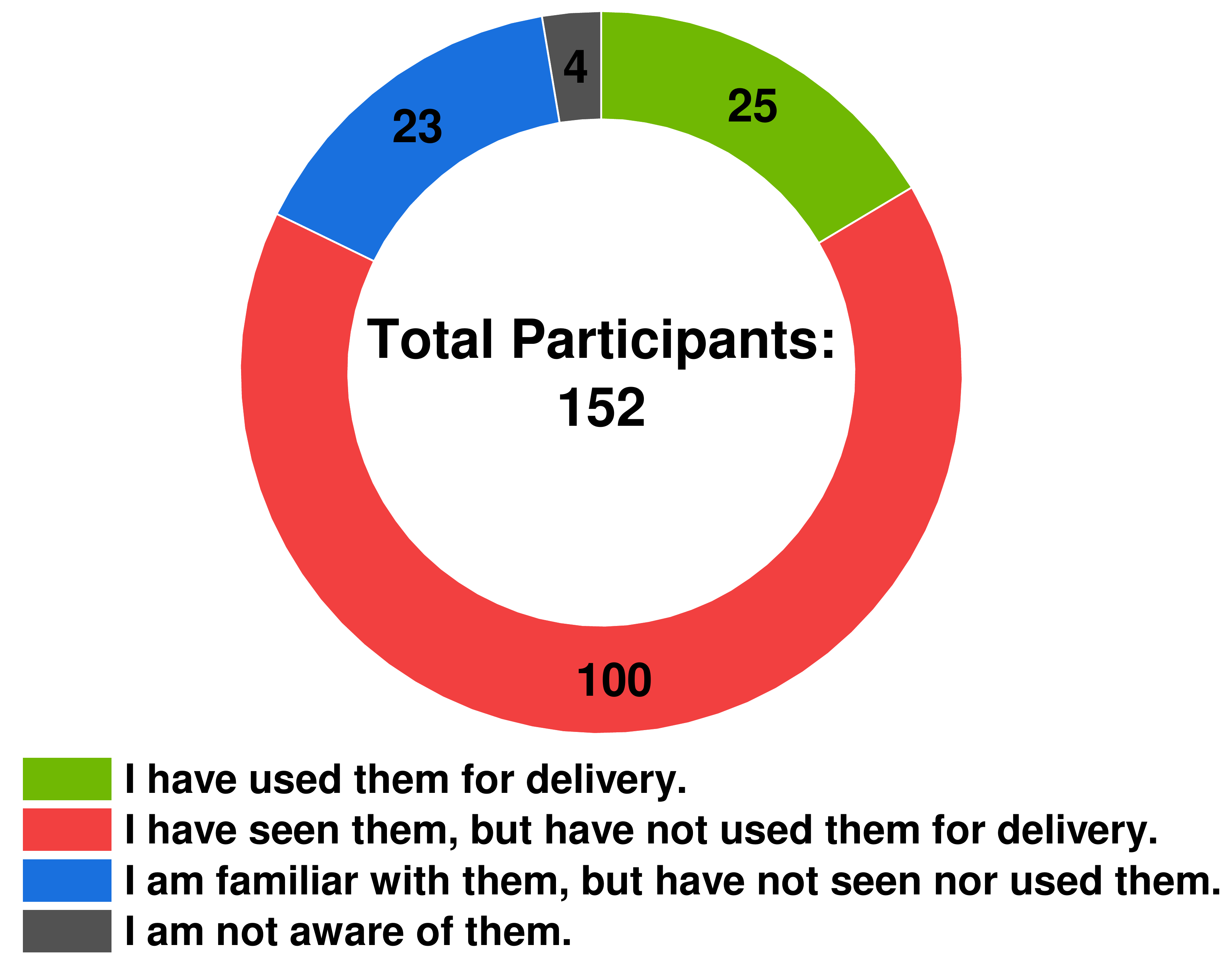}  
  \caption{Distribution of participants' prior experience with delivery robots.}
  \label{fig:delrob_familiar}
\end{subfigure}
\hspace{0.75mm}
\begin{subfigure}{.330\textwidth}
  \centering
  \includegraphics[width=0.99\linewidth]{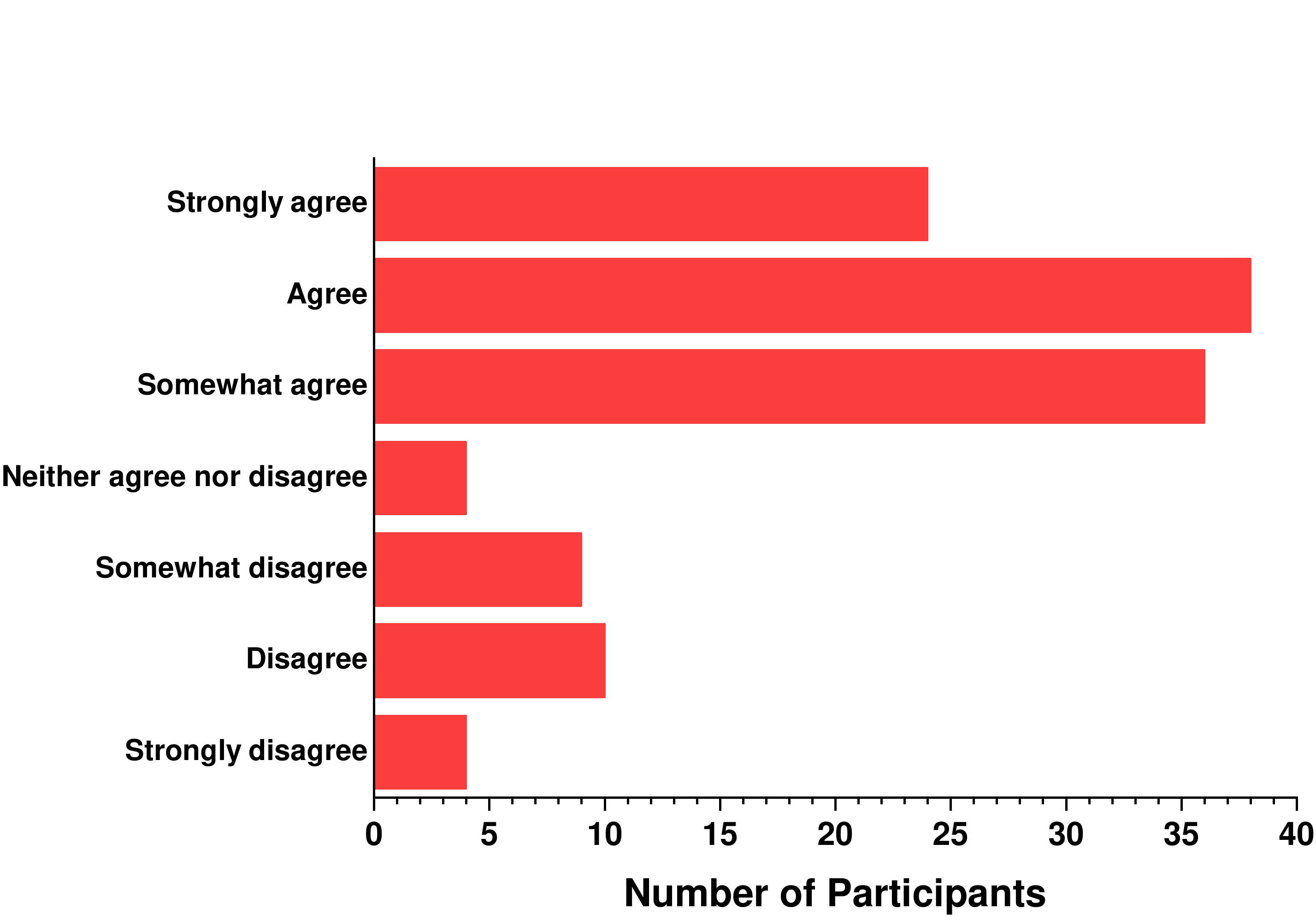}  
  \caption{Distribution of participants' opinion on the uncertainties in delivery robot's navigation.}
  \label{fig:delrob_experience}
\end{subfigure}
\hspace{0.75mm}
\begin{subfigure}{.325\textwidth}
  \centering
  \includegraphics[width=.91\linewidth]{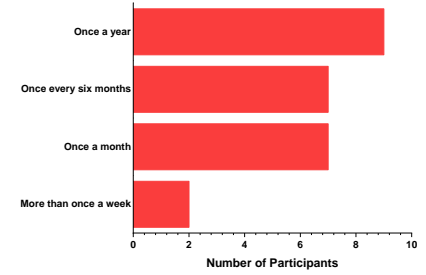}  
  \caption{Distribution of the participants' ordering frequency using delivery robot services.  }
  \label{fig:delrob_order}
\end{subfigure}

\caption{Distribution of the participants' responses to questions on their prior experience with delivery robots and allied services. \textit{Zoom-in for details.}}
\label{fig:participant_demographics}
\vspace{-7mm}
\end{figure*}

\subsection{Background}
Prior researches have investigated on enhancing the predictability of the intents of a mobile robot. It was found that humans perceive about 85\%-90\% of the information through visual means, and visual methods are well suited for conveying the robot intents during human-robot interactions \cite{mangold2015informationspsychologie}. To this end, natural means of communication such as head orientation and arm motions of the robot have been used to express the navigational intent \cite{rosen2020communicating}. However, such methods apply to robots with anthropomorphic features only and are not well suitable for delivery robots in the form of mobile robots that are commonly used.

eHMIs have been widely used as a means for the robot to express their emotion \cite{hong2020multimodal, fitter2016designing, benson2016modeling} and their navigational intent\cite{8520655} for robotics without anthropomorphic features. Out of the various eHMI modalities, LED lights were found to be an efficient means for communicating the robot's intent and have been widely adopted \cite{fernandez2018passive, tafesse2018analysis}. The robots make the best use of various LED colors and patterns to express their intent. Investigations on the use of LED lights on drones in signaling the flight paths found that it largely avoids the uncertainties in the drone's motion and established human's trust in robots \cite{8520655, szafir2015communicating, baraka2016enhancing}. In these prior researches, LED lights in the form of a strip or two lights resembling head lights of a car have been used. However, there has been no consensus on the type of  light nor the light pattern that  is  more  efficient  in  conveying  the  robot’s intent.  Hence,  there  is  a  need  for  further investigation to identify the light type and pattern that is more effective especially for delivery robots.

In addition to lights, display type eHMI modalities are also
widely used to express the robot’s intent \cite{bejerano2018methods}. In \cite{shrestha2016intent}, robot's directional intents were communicated through arrow signs on display and lights, and the efficiency of using lights and display in conveying the intents of the robot was compared. Though the previous researches have found that display is an efficient means of communication, it is presently unknown whether eHMIs should use icon-based graphical content or text-based content to communicate. Icons were found to be more effective, since they do not have language barriers and are more conspicuous \cite{bazilinskyy2019survey}. However, it was also found that interpretability of the icons depends on the prior experience of the humans and text avoids ambiguities \cite{bazilinskyy2019survey}. In addition to the type of content displayed, the inclusion of perspective content (i.e. perspective of the robot) also has been an integral part in the communication. It is currently unclear whether the robot should convey its intent with reference to itself or without any reference to its perspective.

Researchers have also investigated methods to project light-ray and trajectories on the floor that hints at the robots heading direction \cite{matsumaru2006mobile, shrestha2018communicating}. However, a similar system may not be well suited for a delivery robot due to the ever-changing lighting conditions in the outdoors and occlusions.

Based on the current shortcomings, this work focuses on investigating the use of eHMIs for delivery robots. Although numerous types of eHMI modalities like a robotic arm, head motions, sound, display, and light can be used, in this study, the scope is limited to the two most commonly used eHMI modalities: display and light. Display and light were selected as the basic eHMI modality based on its effectiveness in conveying the intent and easy adaptability to delivery robots. For each eHMI modality, two different design elements were considered. Specifically, the display had variations on the type of message displayed (textual and graphical) and with and without the perspective (egocentric versus allocentric \cite{eisma2020external}). For the light, the number of light (single and dual light) and light patterns (flash and sweep pattern) were varied. With these factors, an online survey was conducted to understand the pedestrian preferences of different types of display and light. 


\subsection{Research Question and Hypothesis}
To the best of our knowledge, no prior research on the eHMI design specific to delivery robots has been explored. In this study, we address this research gap with the following research question:

\begin{quote} 
    \textit{\textbf{Q.} Which eHMI modality (display/light) would be most preferred by the pedestrians in understanding the navigational intent of the delivery robot such as forward motion, stop, and turn?}
\end{quote}

An online survey was conducted to explore the user preferences about the content on the eHMI display and the type of light. The following are our research hypothesis:

\begin{itemize}
  \item  \textit{\textbf{H1.1} Graphical content on the eHMI display will be clearer than using textual content to communicate delivery robot's navigational intent. }
  
 \item  \textit{\textbf{H1.2} Content with the perspective of the robot  on the eHMI display will be clearer than content without the perspective of the robot to communicate the delivery robot's intent.}
\end{itemize}
  
 In the display modality, graphical content might be more efficient since they are conspicuous and humans can relate them to the signs they incur in the real world. We also consider that the display contents with the perspective of the robot works better since it does not involve ambiguities such as whether the instruction is for the pedestrian or a message about the robot’s intent.
\begin{itemize}  
  \item \textit{\textbf{H2.1} Dual light is more efficient in conveying the delivery robot intent than single light.}
  
  \item \textit{\textbf{H2.2} Sweeping or continuous light pattern is more efficient in conveying the delivery robot's intent than a flashing light pattern.}
\end{itemize}

In the prior researches, there has been no consensus on the type of light that is more efficient in conveying the robot's intent. Dual lights resembling the lights of the car might align well with the human's prior knowledge and can be understood easily. 
The sweeping pattern of the light might have a similarity with the human's sweeping hand gesture \cite{dey2020color}. Hence, based on Wicken's model of display design \cite{wickens1998introduction}, sweeping pattern might be matched strongly with the humans' mental models and have a better understanding.  

\section{Method}
\subsection{Participants}
The experiment sample consisted of 152 participants who were all over the age of 18. The participants were recruited through various means such as emails, flyers on bulletin boards throughout the university campus, and social media posts. The participants were anonymous, and no identifiable information was collected during the survey. The participants had different prior experiences with delivery robots. Out of the 152 participants, 125 participants (82.2\%) have interacted with delivery robots either for delivery or through observation during a commute on the sidewalk. Delivery robots are currently operational in the authors' university campus for the delivery of food. The majority of the participants were people affiliated with the university, who have seen and have interacted with delivery robots while walking. This made the participants well suited for the research and this study unique since a majority of our participants have already interacted with a delivery robot and are aware of its shortcomings.

The distribution of the participants' prior experience with delivery robots is depicted in Fig.~\ref{fig:delrob_familiar}. The 125 participants who have prior experience interacting with delivery robots (have seen them in public or used them for delivery) were only asked about their opinion on the uncertainties in the delivery robot navigation. On a 7-point Likert scale, 98 participants (78.4\%) agreed that there are ambiguities in the delivery robot navigation. The detailed distribution of the participants' response on the uncertainties on delivery robot navigation is summarized in Fig.~\ref{fig:delrob_experience}. The 25 participants who have used delivery robots for delivery were asked about their ordering frequency and the responses are summarized in Fig.~\ref{fig:delrob_order}.

\subsection{Experimental Design}
The experiment was conducted in a within-subject design. The study focuses on investigating pedestrian preferences on the eHMI design for delivery robots to express their navigational intent. 

\subsubsection{Scenarios}
The experimental design considered three basics scenarios that a pedestrian would experience while interacting with a delivery robot on a sidewalk. The three scenarios were:
\begin{itemize}
    \item \textit{Forward: } A delivery robot keeps moving forward towards the pedestrian. 
    \item \textit{Stop: } A delivery robot in motion comes to a complete stop in front of the pedestrian.
    \item \textit{Turn: } A delivery robot moving in front of the pedestrian makes a left or a right turn.
\end{itemize}

The three scenarios were decided based on the basic actions performed by a mobile robot while navigating. A majority of the robot's navigational tasks can be condensed into these three scenarios. In this study, for the turn scenario left turn was only considered to maintain the brevity of the survey. Also, the results of the left turn can be extrapolated to the right turn as well. 

\subsubsection{eHMI Design Considerations}
In the eHMI design considerations of using display and light systems of delivery robots, further variations were introduced. The independent variables were display with different types of contents (text and graphics) and perspective of the robot on the content (with and without the perspective of the robot). Based on these, the following design variations were considered for the display:

\begin{itemize}
    \item Graphics \textit{w/o} robot perspective: Graphics signs inspired by the signs commonly seen on the road were flashed on the robot display. The graphic designs used did not have any information regarding the context of the robot. 
    \item Graphics \textit{w/} robot perspective: Animated graphic designs using icons of delivery robot were used. The animation conveyed the intent of the robot. The icon of the delivery robot was used to add detail about robot perspective.
    \item Text \textit{w/o} robot perspective: A simple text message conveying the intent of the robot in an action word was flashed on the robot display.
    \item Text \textit{w/} robot perspective: A text message conveying the intent of the robot with the word `ROBOT' appended to the action word. The word `ROBOT' was added to give the readers (pedestrians) a perspective that the action will be performed by the robot.
\end{itemize}

The independent variables within the use of light were a type of light (single and dual light) and light patterns (flash and sweep patterns). Based on these, the following design variations were considered for the light:

\begin{itemize}
    \item Single light with flash pattern: A single long LED strip that flashes colored light was used.
    \item Single light with sweep pattern: A single long LED strip that can be animated to replicate a sweep or a  pattern in various directions was used.
    \item Dual light with flash pattern: Two lights mimicking the headlights of a vehicle that flashes colored light was used.
    \item Dual light with sweep pattern: Two lights mimicking the headlights of a vehicle that can be animated to replicate a sweep or a pattern in various directions was used. 
\end{itemize}

Both the display content and the lights were designed in green, red, or amber based on the scenario. Green was used for the forward scenario, red for the stop scenario, and amber for the turn scenario. The colors were selected based on human's prior knowledge and association of various colors with different actions or intents \cite{bazilinskyy2019survey}. 

\begin{figure}[t]
    \centering
    \includegraphics[width=0.425\textwidth]{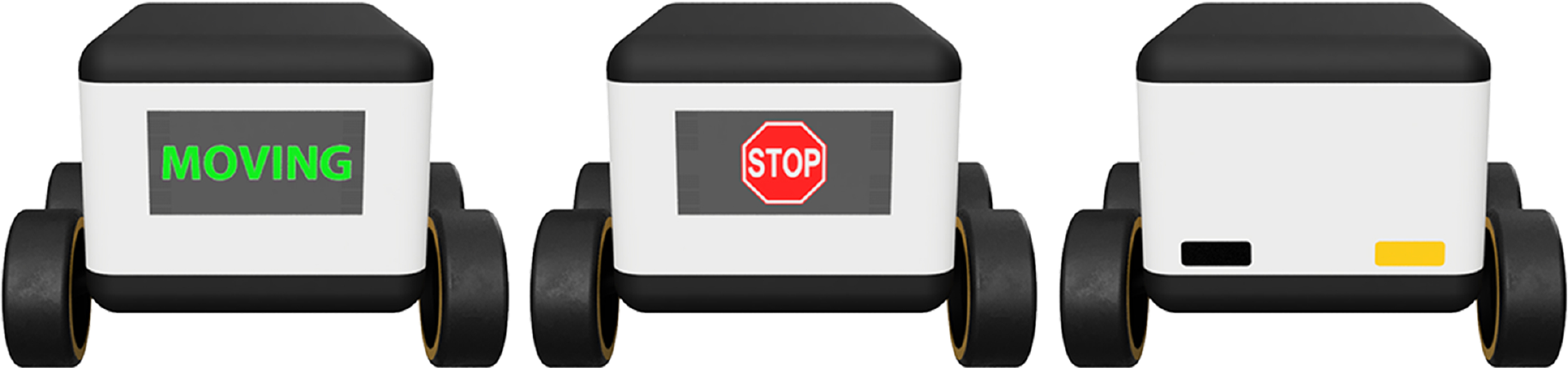}
    \caption{Partial frames from each of the animated designs presented to the participants for each of the three scenarios with text, graphics, and lights (the readers are recommended to refer to the supplementary video for all the animations of the scenarios shown to the participants).}
    \label{fig:robot_animations}
    \vspace{-7mm}
\end{figure}

Animations were created based on the three scenarios and the design considerations. In the scenario animations, a simple delivery robot of size $42\times33\times20$ \si{\cm} whose design was based on commercially available delivery robots with the various eHMIs on its front was made to move in front of a pedestrian. The pedestrian was included to depict the position of the human with respect to the robot. Only a pedestrian and the robot were placed on a plain background to avoid other distractions for the participants. In the animation, the robot moved at a speed of $1.2$ \si{\meter/\second} that is similar to the average walking speed of humans \cite{fitzpatrick2006another}. The animations were created to give the participants a better feel about the scenarios and reduce the gap from a real-world encounter. These animations were shown to the participants during the survey to explain each scenario. Animations based on each design consideration were created with a focus on the front side of the robot. Partial frames from the animation created (one from each scenario) with text, graphics, and lights are shown in Fig.~\ref{fig:robot_animations}. 

\subsection{Experimental Setup}
The survey was conducted through the online survey platform Qualtrics. With mobile internet usage on the rise, it was ensured that the survey was mobile-friendly. The survey consisted of four parts: demographic questions on the participants' prior experience with delivery robots, design evaluation for three scenarios, robot moving, stopping, and turning, respectively. 
The survey began with an age check to ensure that all the participants were over the age of 18. Participants under the age of 18 were not allowed to proceed to the survey. Following the age check, the participants were presented with a description of delivery robots and a purpose of eHMIs on delivery robots. 

The demographic questions inquired on the participants' prior experience with delivery robots. Based on the responses, the participants were asked further questions related to their experience with the delivery robots. The participants who have used delivery robots for delivery service were asked about their ordering frequency. The participants who have interacted with the delivery robot either for delivery or passing by during their commute were asked for their opinion about the ambiguities in the delivery robot navigation. The participant's responses to the demographic questions are summarized in Fig.~\ref{fig:participant_demographics}.

\begin{figure*}[t]
\centering
\begin{subfigure}{.25\textwidth}
  \centering
  \includegraphics[width=.96\linewidth]{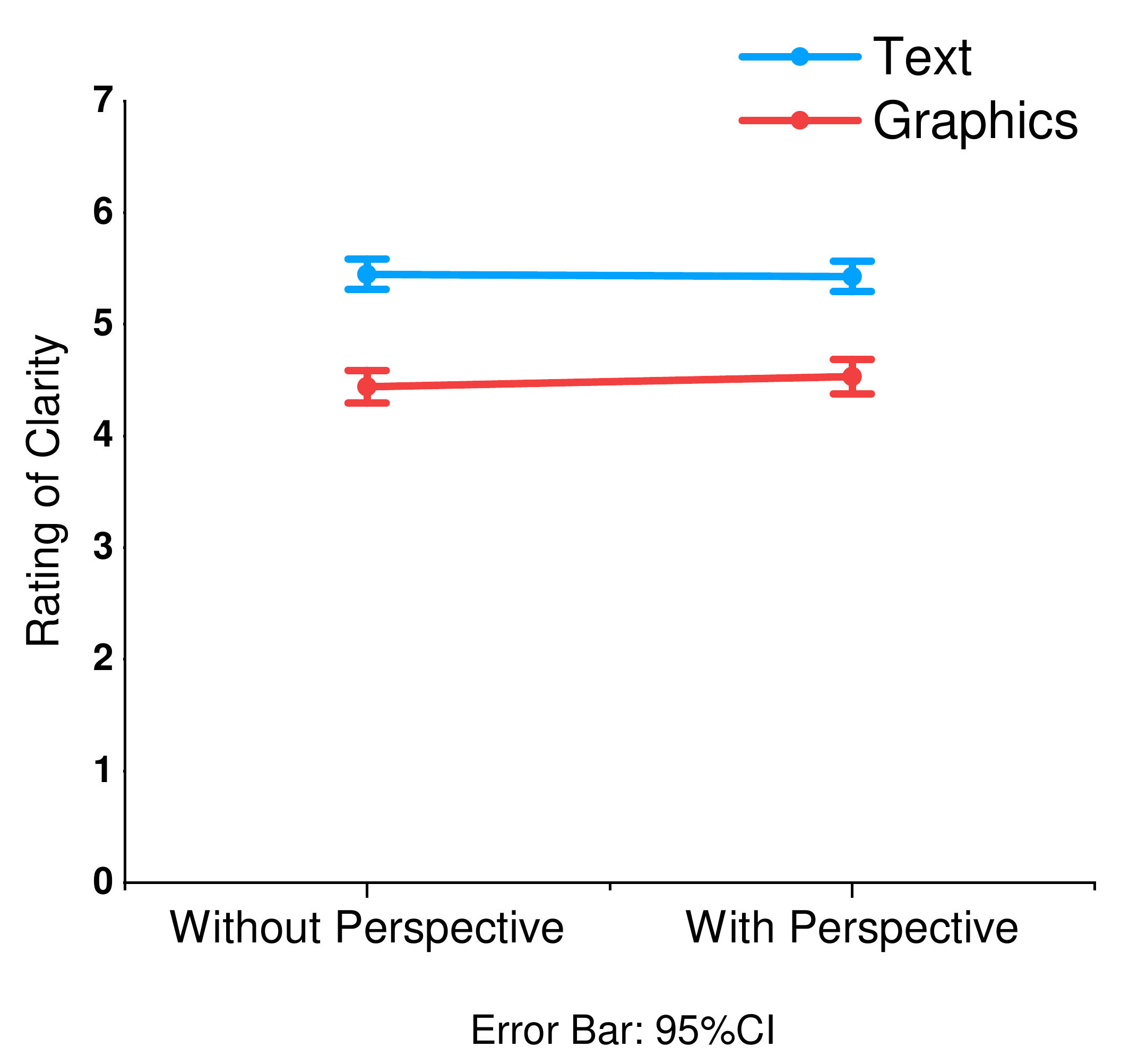}  
  \caption{Forward Scenario}
  \label{fig:forward_display_inter}
\end{subfigure}
\hspace{0.75mm}
\begin{subfigure}{.25\textwidth}
  \centering
  \includegraphics[width=0.96\linewidth]{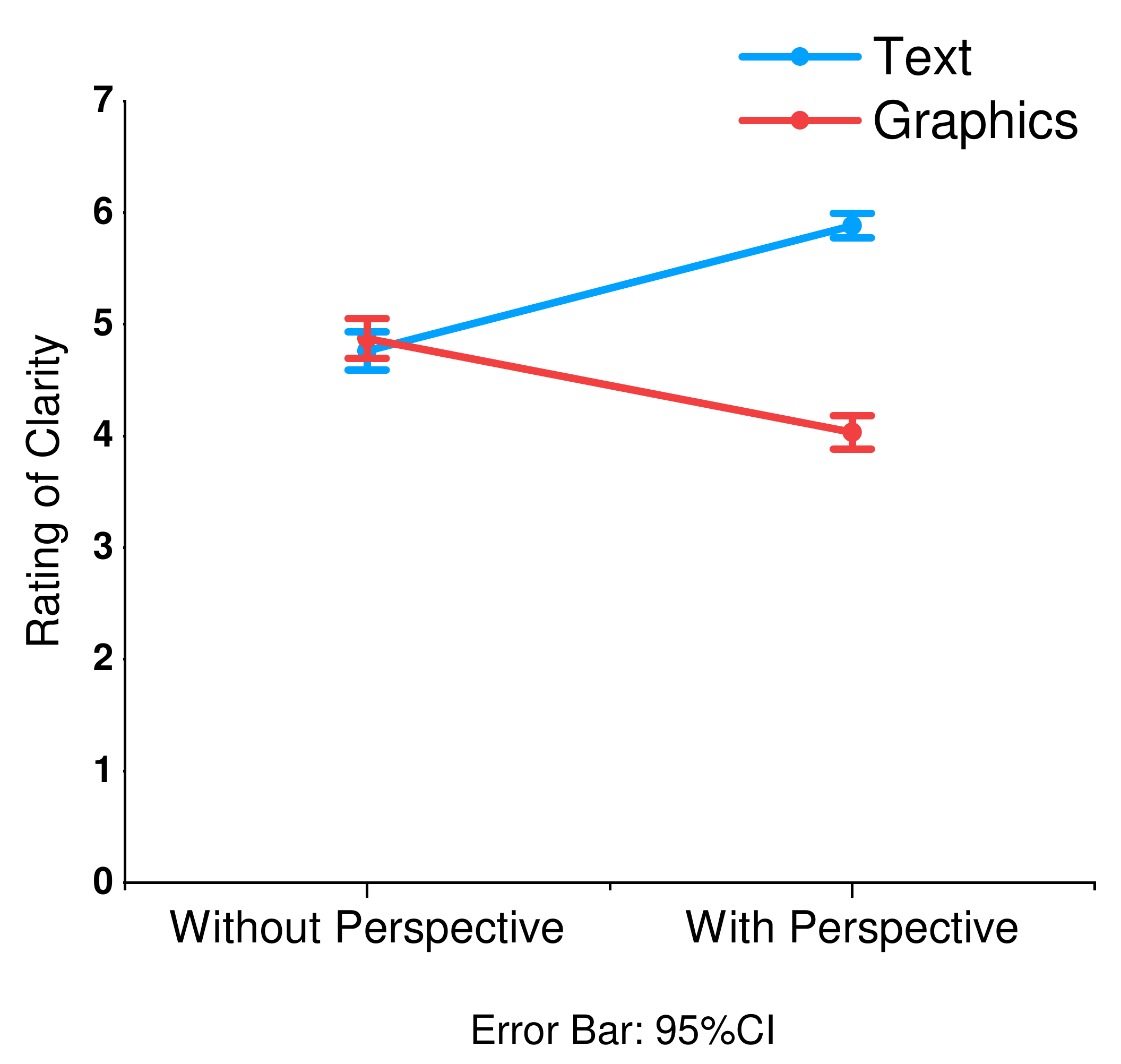}  
  \caption{Stop Scenario}
  \label{fig:stop_display_inter}
\end{subfigure}
\hspace{0.75mm}
\begin{subfigure}{.25\textwidth}
  \centering
  \includegraphics[width=.96\linewidth]{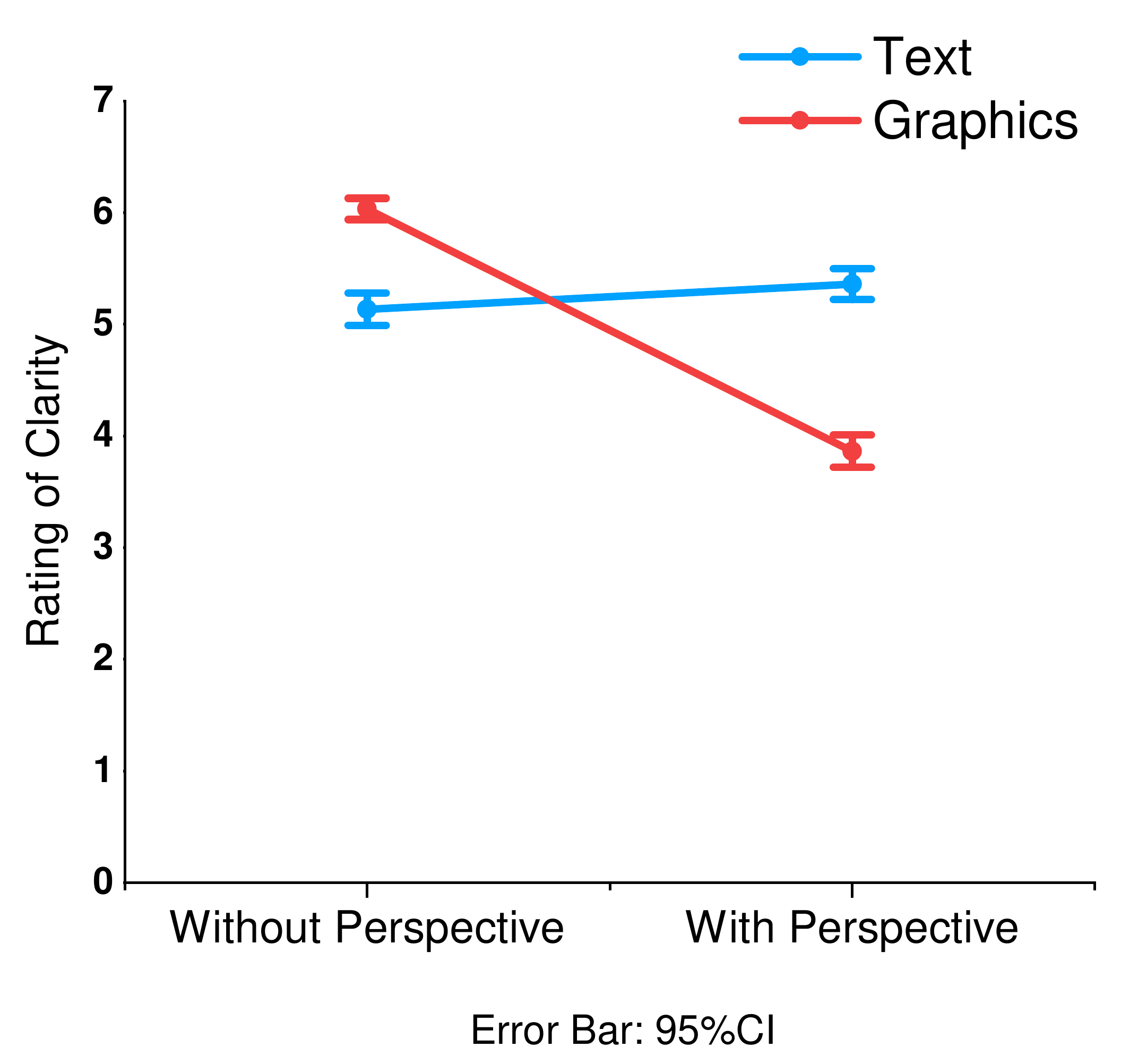}  
  \caption{Turn Scenario}
  \label{fig:turn_display_inter}
\end{subfigure}

\caption{Interaction plots for the variations in the type of display contents and the perspective information about the robot under three different scenarios. Rating of clarity is scored from 1 (Extremely Unclear) to 7 (Extremely Clear).}
\label{fig:interact_display}
\vspace{-4mm}
\end{figure*}

\begin{figure*}[t]
\centering
\begin{subfigure}{.25\textwidth}
  \centering
  \includegraphics[width=.96\linewidth]{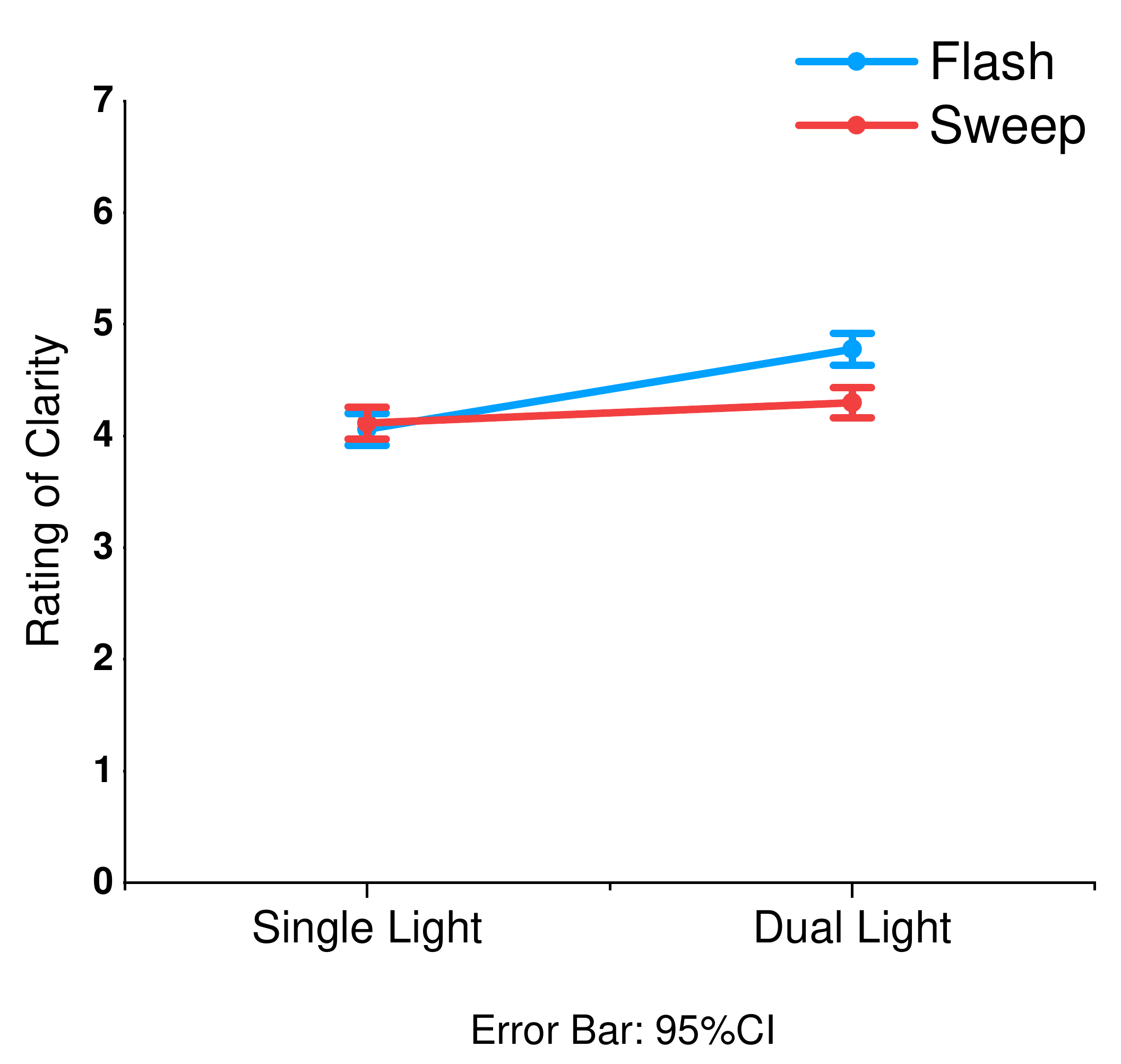}  
  \caption{Forward Scenario}
  \label{fig:forward_light_inter}
\end{subfigure}
\hspace{0.75mm}
\begin{subfigure}{.25\textwidth}
  \centering
  \includegraphics[width=0.96\linewidth]{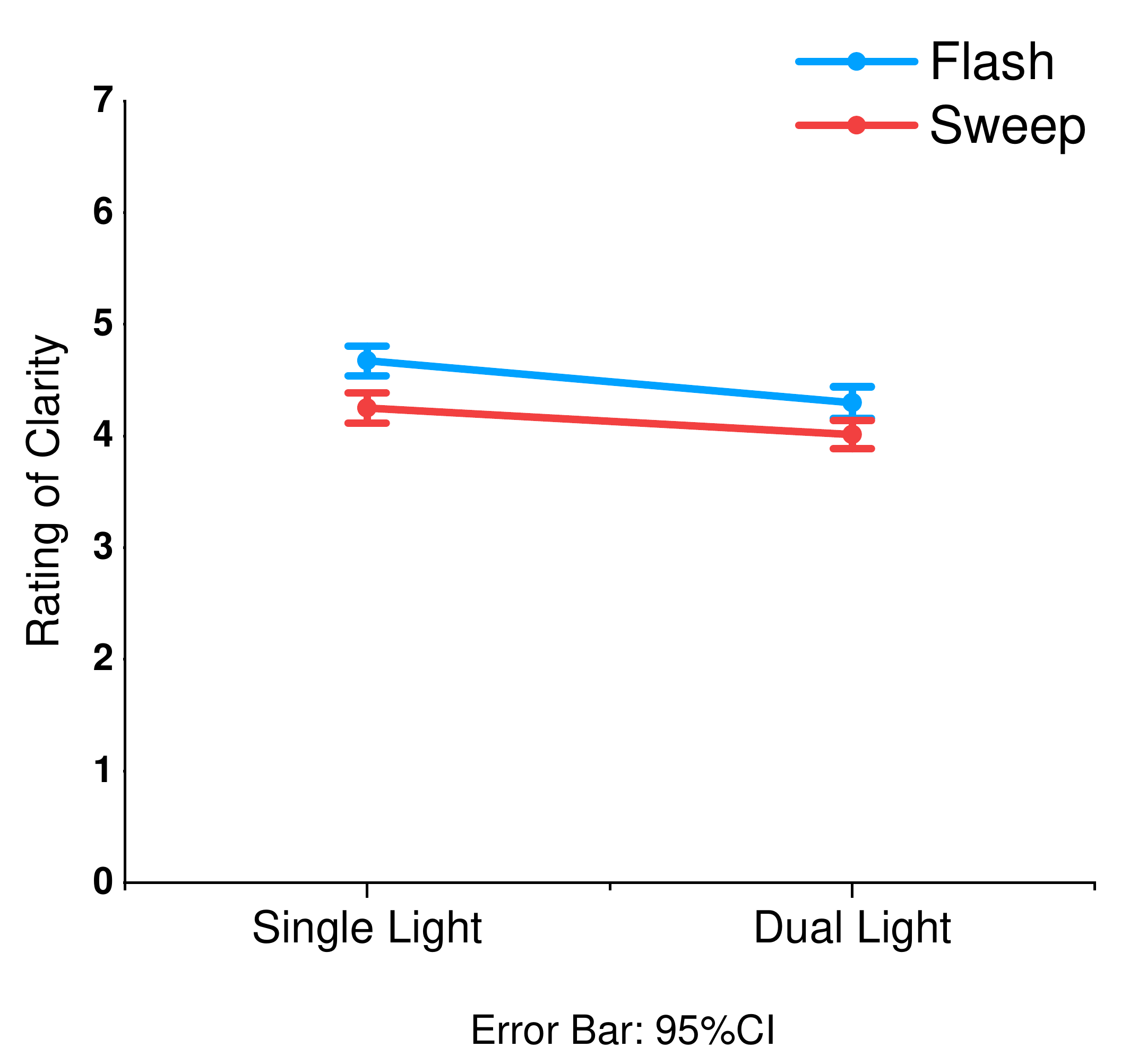}  
  \caption{Stop Scenario}
  \label{fig:stop_light_inter}
\end{subfigure}
\hspace{0.75mm}
\begin{subfigure}{.25\textwidth}
  \centering
  \includegraphics[width=.96\linewidth]{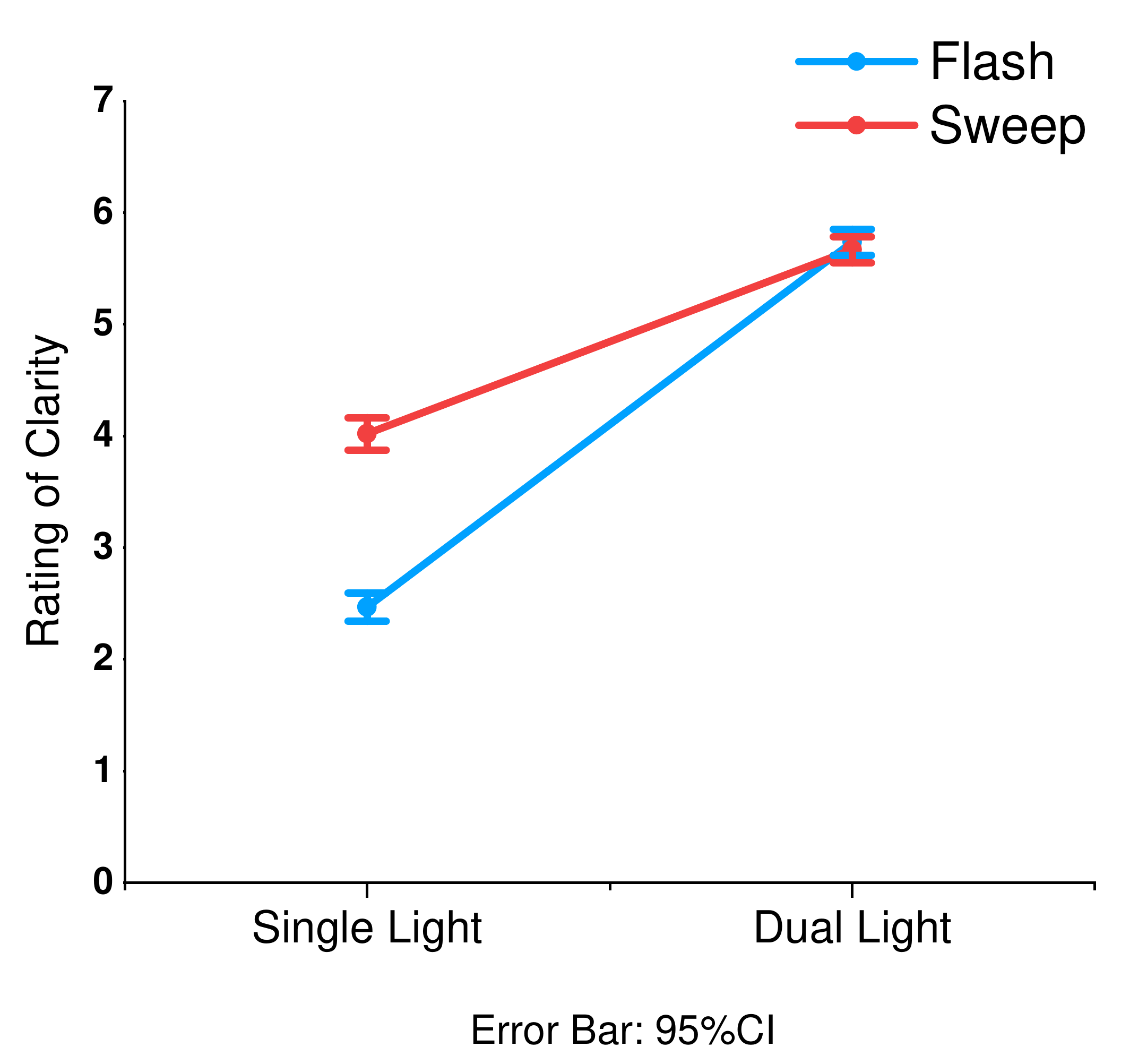}  
  \caption{Turn Scenario}
  \label{fig:turn_light_inter}
\end{subfigure}

\caption{Interaction plots for the variations in the type of light and different light patterns under three different scenarios. Rating of clarity is scored from 1 (Extremely Unclear) to 7 (Extremely Clear).}
\label{fig:interact_light}
\vspace{-8mm}
\end{figure*}

With three scenarios and eight eHMI designs in each scenario, the participants experienced 24 eHMI designs during the survey. The survey consisted of three pages, one page per scenario. The eHMI designs were grouped based on the scenarios to make it effortless for the participants to be aware of the scenario. Each page of the survey elaborated the context of the scenario using text and animated videos similar to the setup shown in Fig.~\ref{fig:intro_figure}.

Following, the explanation of the scenario the participants were presented with the animations of eHMI designs corresponding to that scenario in random order. The participants were asked to rate the eHMI design from the perspective of a pedestrian based on the design's clarity on explaining the delivery robot's intent. The participants rated the designs on a 7-Likert scale, with 7 implying that the design is `Extremely Clear' and 1 implying that the design is `Extremely Unclear'. The rating of 4 implies that the design is `Neither Clear nor Unclear'. Here, `Extremely Clear' implies that the scenario and the robot's intent expression are well related and the expression methods signals the intent of the robot without any ambiguity. Once, the eight designs were presented for rating, the participants were asked to rank the designs with rank 1 for the most preferred design and rank 8 for the least preferred one. A drag and drop setup were provided to enable easy ranking of the designs. In case the participants had an equal preference for two or more designs, the participants were asked to convey the same on a free-form text box that followed the ranking question.

The participants were then asked about their opinion on using both light and the display to express the delivery robot's intent. The participants answered a binary question, implying whether they feel the combination enhances the clarity of the robot's intent or not. If the participant implied that the combination of display and light improves the clarity, they were asked about their preferred combination. The participants picked one light-based design and one display-based design. At the end of each scenario, the participants were asked for their subjective opinion on the eHMI design for delivery robots using a free-form text box. The same setup was followed for all three scenarios. The scenarios were presented to the participants in the following order: moving forward, stopping, and turning. 

Prior to the deployment, a pilot study was conducted to validate the clarity of the instructions. The participants took 6 to 8 minutes to complete the survey. The completion times were recorded to set a baseline on the average time required to complete the survey. 

\section{Results}
\subsection{Data Sanity Check}
With the collected data, we first filtered out invalid responses. 
In crowdsourced data collection, the participants tend to respond by skipping the instructions. Hence, they took less time to complete the survey. Out of the 152 participants, five participants took less than 5 minutes to complete the survey. The time taken for the completion was way less than the baseline completion time of 6 to 8 minutes. Those responses were not analyzed and this led to a slightly reduced sample size of 147.

\subsection{Results of Clarity Rating}
Two-way repeated-measures ANOVA tests were conducted using the rating of the participants with the 7-point Likert Scale on all three scenarios separately. The tests were carried out to investigate the effect of various eHMI designs on the participants' ratings about the eHMIs clarity. Mauchly's test for sphericity was conducted and the assumption of sphericity was not violated.

\subsubsection{Display-based eHMIs}
The type of content on the display (text or graphics) and perspective of the robot on the content were used as the independent variables, and the rating of the participants was used as the dependent variable for data analysis. For the forward scenario, the results showed that there was a significant change in the ratings of the participants when the type of contents on the display was changed ($F=36.24$, $p< 0.001$, $\eta^2=0.199$). Textual content ($\mu=5.8$) had a higher rating compared to graphics ($\mu=4.74$). But, there were no significant changes in the ratings of the participants whether the content of the display has the perspective information of the robot or not ($F=0.98$, $p=0.754$, $\eta^2=0.001$). The interaction between the type of content and the perspective information was not significant (see Fig.~\ref{fig:forward_display_inter}). 

For the stop scenario, a significant interaction was found between independent variables considered for the design of display eHMI  ($F=94.282$, $p< 0.001$, $\eta^2=0.392$) (see Fig.~\ref{fig:stop_display_inter}). Simple main effect analysis showed that both content types on display (text and graphics) had a statically significant difference on the rating (text: $\mu=5.32$; graphics: $\mu=4.45$; $p< 0.001$). When there was no perspective information on the display content no difference was found ($p=0.28$), but difference in the ratings were found when the perspective information was introduced ($p< 0.001$).

Lastly, for the turn scenario, a significant interaction effect between the variables was derived ($F=207.482$, $p<0.001 $, $\eta^2=0.587$) (see Fig.~\ref{fig:turn_display_inter}). Simple main effect analysis on the turn scenario showed that textual content did not have a significant difference ($p=0.41$), whereas for the graphical content there was a significant difference, with graphical content without perspective ($\mu=5.361$) rated higher over the ones with perspective ($\mu=3.864$). 

\subsubsection{Light-based eHMIs}
To evaluate the influence of the light-based eHMI, the type of light (single and dual) and the light patterns (flash and sweep) were used as the independent variables, and the rating of the participants was used as the dependent variable. In the forward-moving scenario, a significant interaction effect was found ($F=12.217$, $p=0.001 $, $\eta^2=0.077$). This interaction effect between the variables when the robot is moving forward is shown in (Fig.~\ref{fig:forward_light_inter}). Results of simple main effect analysis showed flashing pattern of light had a significant impact on the rating ($p< 0.001$), whereas the sweep pattern did not ($p=0.225$). For the flashing pattern, dual lights were preferred to the single lights ($\mu=4.776$ \textit{vs} $\mu=4.016$). Similarly, simple main effect analysis based on types of lights showed that single light did not make any difference  ($p=0.645$), whereas dual lights did ($p< 0.001$). With the dual lights, flashing pattern ($\mu=4.776$) was more comprehensible over the sweep pattern ($\mu=4.299$). 

The tests on the ratings from the stop scenario showed that participants significantly preferred the flashing patterns to the sweeping patterns ($F=12.792$, $p<0.001 $, $\eta^2=0.081$). There was also a significant difference in the rating when different light patterns were used for the stop scenario ($F=16.120$, $p<0.001 $, $\eta^2=0.081$). The interaction effect was not significant in this case (see \ref{fig:stop_light_inter}).

Similar evaluations on the participants' rating on the turn scenario showed a statistically significant interaction effect between the use of different types of lights and light patterns ($F=85.935$, $p<0.001 $, $\eta^2=0.371$). This interaction effect is depicted in Fig.~\ref{fig:turn_light_inter}. In the simple main effect analysis, both flash ($p< 0.001$) and sweep ($p< 0.001$) patterns created significant difference in the ratings. For both the flash and sweep patterns, dual lights (flash: $\mu=5.735$; sweep: $\mu=5.584$ ) were highly preferred over single (flash: $\mu=2.469$; sweep:$\mu=4.020$ ). 
In the analysis on single and dual light, single lights ($p< 0.001$) had significant difference on the rating, whereas dual lights ($p=0.359$) did not. Within the single light strip, sweep pattern ($\mu=4.020$) was rated high over the flash pattern ($\mu=2.469$). 

\begin{table*}
\centering
\caption{Rank responses of the participants for each scenario. The most preferred design (rank 1) was given a normalized score of 1 and the least preferred design (rank 8) was given a normalized score of 0.}
\vspace{-1mm}
\scriptsize
\label{tab:rank_average}
\begin{tabular}{cccclcclcclcc} 
\toprule
\multicolumn{1}{l}{}     & \multicolumn{1}{l}{} & \multicolumn{5}{c}{Light}                                        &  & \multicolumn{5}{c}{Display}                                             \\ 
\cline{3-7}\cline{9-13}
                         &                      & \multicolumn{2}{c}{Single LED} &  & \multicolumn{2}{c}{Two LEDs} &  & \multicolumn{2}{c}{Graphic}      &  & \multicolumn{2}{c}{Text}          \\ 
\cline{3-4}\cline{6-7}\cline{9-10}\cline{12-13}
                         &                      & Flash       & Sweep       &  & Flash       & Sweep     &  & W/O Persp. & W/ Persp. &  & W/O Persp. & W/ Persp.  \\ 
\cline{3-7} \cline{9-13}
\multirow{2}{*}{Forward} & Mean                 & 0.445 & 0.681      &  & 0.465 & 0.555    &  & 0.526     & 0.498     &  & 0.732     & 0.664     \\
                         & S.D.                 & 0.250 & 0.213      &  & 0.244 & 0.238  &  & 0.294     & 0.324    &  & 0.259     & 0.310     \\
\multirow{2}{*}{Stop}    & Mean                 & 0.636 & 0.636      &  & 0.394 & 0.418    &  & 0.694     & 0.421    &  & 0.643     & 0.761     \\
                         & S.D.                 & 0.221 & 0.208       &  & 0.263  & 0.185    &  & 0.331     & 0.281    &  & 0.278     & 0.235      \\
\multirow{2}{*}{Turn}    & Mean                 & 0.298 & 0.482      &  & 0.657 & 0.651    &  & 0.857     & 0.390    &  & 0.625           & 0.599     \\
                         & S.D.                 & 0.214 & 0.246      &  & 0.279 & 0.251    &  & 0.165     & 0.234    &  & 0.256     & 0.250     \\
\bottomrule
\end{tabular}
\vspace{-3mm}
\end{table*}

\begin{table*}
\centering
\caption{Frequency distribution of the participants' preferences for various display and light combinations under the three different scenarios.}
\vspace{-1mm}
\scriptsize
\label{tab:combinations}
\begin{tabular}{llcccc} 
\toprule\toprule
Scenario                 & \multicolumn{1}{c}{\diagbox{Display}{Light}} & Single Light/Flash~ & Single Light/Sweep & Dual Light/Flash~ & Dual Light/Sweep~  \\
\hline\hline
\multirow{4}{*}{Forward} & Text w/o~ perspective                        & 13                  & 8                  & 7                  & 3                   \\
                         & Text w/~ perspective                         & 5                   & 13                 & 5                  & 4                   \\
                         & Graphics w/o~ perspective                    & 5                   & 11                 & 5                  & 3                   \\
                         & Graphics w/~ perspective                     & 4                   & 5                  & 2                  & 4                   \\
                         \hline
\multirow{4}{*}{Stop}    & Text w/o~ perspective                        & 3                   & 2                  & 2                  & 1                   \\
                         & Text w/~ perspective                         & 14                  & 13                 & 10                 & 3                   \\
                         & Graphics w/o~ perspective                    & 11                  & 2                  & 7                  & 2                   \\
                         & Graphics w/~ perspective                     & 3                   & 0                  & 2                  & 3                   \\
                         \hline
\multirow{4}{*}{Turn}    & Text w/o~ perspective                        & 3                   & 3                  & 3                  & 2                   \\
                         & Text w/~ perspective                         & 3                   & 5                  & 4                  & 4                   \\
                         & Graphics w/o~ perspective                    & 0                   & 2                  & 1                  & 1                   \\
                         & Graphics w/~ perspective                     & 15                  & 7                  & 11                 & 4                   \\
\bottomrule\bottomrule
\end{tabular}
\vspace{-7mm}
\end{table*}

\subsection{Results of Ranking}
The rank responses of the participants were analyzed to find the most preferred eHMI design for each scenario. The rank data was based on the order of the participants' design preferences and responses specified on the text box, in case of equal preference for two or more designs. In order to analyze the rank data, the participants' responses were normalized to be between 0 and 1. The most preferred design (rank 1) was given a normalized score of 1, and the least preferred design (rank 8) was given a normalized score of 0. The other ranks were allocated scores between 0 and 1. The normalized scores were then averaged to compute a preference metric for each design. The means of the normalized scores along with the standard deviations for all the eHMI designs under the three scenarios are presented in Table \ref{tab:rank_average}. 

Based on the means of the normalized scores, display with text without any perspective is most preferred when the robot is moving forward ($\mu = 0.732$ and $\sigma=0.259$). Display with text and perspective, and single LED with the sweeping pattern are also highly preferred. For the stop scenario, the display with text and the perspective is the most preferred design ($\mu=0.762$ and $\sigma=0.235$), followed by
the display with graphic but without perspective ($\mu=0.694$ and $\sigma=0.331$).
Finally, for the turn scenario, display with sign is the most preferred one ($\mu = 0.857$ and $\sigma=0.165$).


\subsection{Results of Combination Preferences}
Out of the 147 responses, 97 participants indicated that they prefer a combination of display and light for the forward scenario. Similarly, 78 participants indicated that they prefer a combination for the stop scenario and 68 participants preferred a combination for the turn scenario. The frequency distribution of the participants' preferences for various display and light combinations is summarized in Table \ref{tab:combinations}. A Chi-square test was performed on the data to find any significant relationship between the two eHMI modalities.  A statistically significant relationship was not found for forward ($\chi^2=9.65, p=0.3796$), stop ($\chi^2=11.97, p=0.215$), nor the turn ($\chi^2=6.74, p=0.6642$) scenarios. Hence, though a combination of display and light was preferred no strong relationship exists between them.

\subsection{Qualitative Data}
The responses of the participants to the open-ended qualitative question at the end of each scenario were analyzed thematically. The insights from the participants' comments are presented along with supporting quotes. The responses made by the participants while rating the forward scenario designs are prefixed with the character `F'. Similarly, the responses under the stop and turn scenarios are prefixed with the character `S' and `T', respectively.

\subsubsection{Universal Designs} Though the use of text on the display to express the delivery robot's intent was clear under different scenarios, the participants concerned that not all people can understand the textual content due to language barriers. 

\noindent (T15) \textit{``Using text like `left' or `right' may create accessibility issues for people who cannot read or do not know English." }
\noindent (F105) \textit{``Not everyone can read. Small kids can learn quicker with signals." }
\noindent (S12) \textit{``The word Stop is good, although I like the picture of the robot with an X since it is understandable in any language."}

\subsubsection{Auditory Feedback} In addition to the display and light, the use of audio either non-verbal or verbal feedback to alert the pedestrian adds value to the design.

\noindent (F40) \textit{``Mild sound signal can be added as one more signalling method which will alert the human close by the robot." }

\noindent (F133) \textit{``Some greeting beeps to alert pedestrians that are not paying attention (texting, talking to a friend, etc.). I'm thinking a R2D2 style greeting."} 

\subsection{Evaluation of Hypothesis}
Based on the data and the results, the hypothesis was validated as follows:
\begin{itemize}
  \item  \textit{H1.1 Graphical content on the eHMI display will be not be clearer than textual content for communicating the robot's intent while moving forward.}
\end{itemize}

  The quantitative data from the statistical analysis and the rank information supports the same. 
\begin{itemize}  
  \item \textit{H1.2 Content with the perspective of the robot on the eHMI display will be not clearer than content without the perspective of the robot to communicate the delivery robot's intent when the robot is moving forward.} 
\end{itemize}
 The statistical results revealed that there was no significant difference in the results when display with and without perspective information was used. But, a strong conclusion could not be made for the stop and the turn scenarios due to the interaction effects. Rather than that, we found that preference for textual to graphical content was only happened when the information included the perspective of the robot in stop and turn scenarios. 
\begin{itemize}  
  \item \textit{H2.1 Dual light is not more efficient in conveying the delivery robot intent than single light for stop scenario.} 
\end{itemize}
  
  Though the statistical results from the stop scenario showed that there was a significant difference in the results for the use two different types of lights, the means of the rating from Table \ref{tab:rank_average} showed that single light was preferred over dual light. 
\begin{itemize}
  \item \textit{H2.2 Sweeping or continuous light pattern is more efficient in conveying the delivery robot's intent than a flashing light pattern.}
\end{itemize}
  The statistical analysis concludes that there was a difference in the results between the use of sweeping and flashing light patterns in the stop scenario. Again, a strong conclusion could not made for the forward and the turn scenarios due to the interaction effects. 

\section{Discussion}
In this work, a survey was conducted to find the effect of display and light-based eHMI in conveying the navigational intent of the delivery robot to the pedestrians, and the results were summarized. 

\noindent\textbf{Implications of Display.~}
The participants (F68, F73, F118, S58, S68, S69, S103, S122, T101, T129) qualitatively indicated suggestions or alternatives to the text. From the comments, it was seen that the using present particles in the text to convey the robot's intent will be clearer. The use of present particle clarifies the perceptive information too without the use of the `ROBOT' word. For instance, the use of the word `STOP' was misleading. The participants assumed the robot is instructing the pedestrian to stop rather than conveying its intent. 

The participants preferences over display as the primary eHMI modality opens up many gates for further research. The participants indicated that certain textual messages were clear in understanding the intent and some were not. Studies on people's understanding of various words related to robot's navigation might be needed in identifying the universally understood words to express robot's intent. Mixed results from rating and ranking data were observed between the use of textual and graphical contents on the display. For instance, left arrow sign (graphical design without robot perspective) was highly preferred to indicate robots turning intent. But robot icon (graphical design with robot perspective) was highly preferred for the robot stop case. This indicates that even with one form of eHMI one independent variable cannot be used for the robot actions. Different independent variables are good at conveying different intents of the robots. 

\noindent\textbf{Implications of Light.~}
Though display$(\mu=4.922)$ was preferred over light $(\mu=4.630)$ as a primary eHMI modality, participants (F11, F39, T23) qualitatively expressed that light as a redundant information in addition to the display was preferred under all the scenarios. Commercially lights are not expense, it should be feasible to add light as a redundant information. Some participants (F39, F41, F68) preferred having eHMIs on multiple sides of the robot rather than just at the front. Adding lights on all sides of the design might be a commercially viable design over having multiple displays. 

Based on the ratings for the stop scenario, the use of single light was preferred over the use of display. However, display was preferred for the other two scenarios. Under such cases, the light patterns can be mimicked from the display rather than adding lights in addition to the display just for one scenario. The human understanding and preference on creating light patterns from the display needs further investigation before actual implementation. 

\noindent\textbf{Implications on Combinations.~}
Though a relationship between different display and light types could not be established statistically, the interesting observation were made on the responses. Out of the 147 participants, 97 indicated that they prefer a combination to indicate the robot forward scenario. Opposite results were seen for the robot turn scenarios. This shows that certain intents are better shown with combinations, whereas certain scenarios are fine with one. Even in the participants responses for the combination preference for the turn, the left arrow sign was selected widely. The participants preferred that sign not only when used individually but also when used along the light. Similar pattern was observed in the stop scenario too. Indicating robot stop intent using text with the perspective of the robot was selected by over 50\% of the respondents under both cases. 

Further observations on the responses to combinations indicated that the single light was preferred over dual light when used with any display content. This contracts with the results of individual preferences where dual lights were preferred over single light. Based on this it suggests that the eHMI modality that is preferred the most individually may not be preferred when used with another eHMI device. Hence, the inferences on the eHMI preferences cannot be transferred from one setup to another. In real world, this implies that the design cues of a delivery robot may not suit well for another one. The robots need to be designed based on a wholesome evaluation rather than drawing inspirations from other robots. 

\noindent\textbf{Effect of Scenarios.~}
The participants preferred combined designs for forward and the stop scenarios. Nevertheless, most of the participants were fine with single eHMI for the turn scenario. Although displaying graphical sign without any robot perspective was most preferred, other contents on the display and dual light designs were clear too. This may be due to the fact that humans are so used to the turn scenario from the vehicles. Vehicles do not signal forward motion with any external cues. Because of that, the robot expressing its intent of moving forward may be something unfamiliar to the humans. Using flashing lights when a brake is pressed in common in vehicles. However, there was not any inclination of the participants' responses to flashing lights for the stop scenario. 

\section{Conclusion}
In this paper, an online crowd-source based evaluation of using display and light as external human-machine interface on a delivery robot to express its navigational intent to the pedestrian has been explored. The results showed that display was effective in conveying the intent of the robot and that a combination of display and light was also preferred based on the scenario due to the presence of redundant information. Also, the results indicated that certain eHMI designs were well suited for specific scenarios and hence, adapting eHMI modality based on the scenario would be the most suitable one. The insights on the effect of various eHMI designs explored in this paper can be used for the advancement of delivery robots where they can communicate with humans for a smooth navigation. 

\bibliographystyle{IEEEtran}
\bibliography{references}

\begin{thebibliography}{10}
\providecommand{\url}[1]{#1}
\csname url@rmstyle\endcsname
\providecommand{\newblock}{\relax}
\providecommand{\bibinfo}[2]{#2}
\providecommand\BIBentrySTDinterwordspacing{\spaceskip=0pt\relax}
\providecommand\BIBentryALTinterwordstretchfactor{4}
\providecommand\BIBentryALTinterwordspacing{\spaceskip=\fontdimen2\font plus
\BIBentryALTinterwordstretchfactor\fontdimen3\font minus
  \fontdimen4\font\relax}
\providecommand\BIBforeignlanguage[2]{{%
\expandafter\ifx\csname l@#1\endcsname\relax
\typeout{** WARNING: IEEEtran.bst: No hyphenation pattern has been}%
\typeout{** loaded for the language `#1'. Using the pattern for}%
\typeout{** the default language instead.}%
\else
\language=\csname l@#1\endcsname
\fi
#2}}

\bibitem{wood2015interpersonal}
J.~T. Wood, \emph{Interpersonal communication: Everyday encounters}.\hskip 1em
  plus 0.5em minus 0.4em\relax Nelson Education, 2015.

\bibitem{birdwhistell2010kinesics}
R.~L. Birdwhistell, \emph{Kinesics and context: Essays on body motion
  communication}.\hskip 1em plus 0.5em minus 0.4em\relax University of
  Pennsylvania press, 2010.

\bibitem{matsumaru2006mobile}
T.~Matsumaru, T.~Kusada, and K.~Iwase, ``Mobile robot with
  preliminary-announcement function of forthcoming motion using light-ray,'' in
  \emph{2006 IEEE/RSJ International Conference on Intelligent Robots and
  Systems}.\hskip 1em plus 0.5em minus 0.4em\relax IEEE, 2006, pp. 1516--1523.

\bibitem{shrestha2018communicating}
M.~C. Shrestha, T.~Onishi, A.~Kobayashi, M.~Kamezaki, and S.~Sugano,
  ``Communicating directional intent in robot navigation using projection
  indicators,'' in \emph{2018 27th IEEE International Symposium on Robot and
  Human Interactive Communication (RO-MAN)}.\hskip 1em plus 0.5em minus
  0.4em\relax IEEE, 2018, pp. 746--751.

\bibitem{fernandez2018passive}
R.~Fernandez, N.~John, S.~Kirmani, J.~Hart, J.~Sinapov, and P.~Stone, ``Passive
  demonstrations of light-based robot signals for improved human
  interpretability,'' in \emph{2018 27th IEEE International Symposium on Robot
  and Human Interactive Communication (RO-MAN)}.\hskip 1em plus 0.5em minus
  0.4em\relax IEEE, 2018, pp. 234--239.

\bibitem{tafesse2018analysis}
Y.~D. Tafesse, M.~Wigness, and J.~Twigg, ``Analysis techniques for displaying
  robot intent with led patterns,'' US Army Research Laboratory Adelphi, Tech.
  Rep., 2018.

\bibitem{watanabe2015communicating}
A.~Watanabe, T.~Ikeda, Y.~Morales, K.~Shinozawa, T.~Miyashita, and N.~Hagita,
  ``Communicating robotic navigational intentions,'' in \emph{2015 IEEE/RSJ
  International Conference on Intelligent Robots and Systems (IROS)}.\hskip 1em
  plus 0.5em minus 0.4em\relax IEEE, 2015, pp. 5763--5769.

\bibitem{baraka2016enhancing}
K.~Baraka, S.~Rosenthal, and M.~Veloso, ``Enhancing human understanding of a
  mobile robot's state and actions using expressive lights,'' in \emph{2016
  25th IEEE International Symposium on Robot and Human Interactive
  Communication (RO-MAN)}.\hskip 1em plus 0.5em minus 0.4em\relax IEEE, 2016,
  pp. 652--657.

\bibitem{trautman2010unfreezing}
P.~Trautman and A.~Krause, ``Unfreezing the robot: Navigation in dense,
  interacting crowds,'' in \emph{2010 IEEE/RSJ International Conference on
  Intelligent Robots and Systems}.\hskip 1em plus 0.5em minus 0.4em\relax IEEE,
  2010, pp. 797--803.

\bibitem{mangold2015informationspsychologie}
R.~Mangold, \emph{Informationspsychologie: Wahrnehmen und gestalten in der
  medienwelt}.\hskip 1em plus 0.5em minus 0.4em\relax Springer-Verlag, 2015.

\bibitem{rosen2020communicating}
E.~Rosen, D.~Whitney, E.~Phillips, G.~Chien, J.~Tompkin, G.~Konidaris, and
  S.~Tellex, ``Communicating robot arm motion intent through mixed reality
  head-mounted displays,'' in \emph{Robotics research}.\hskip 1em plus 0.5em
  minus 0.4em\relax Springer, 2020, pp. 301--316.

\bibitem{hong2020multimodal}
A.~Hong, N.~Lunscher, T.~Hu, Y.~Tsuboi, X.~Zhang, S.~F. dos Reis~Alves,
  G.~Nejat, and B.~Benhabib, ``A multimodal emotional human-robot interaction
  architecture for social robots engaged in bidirectional
  communication$\backslash$vspace* 7pt,'' \emph{IEEE transactions on
  cybernetics}, 2020.

\bibitem{fitter2016designing}
N.~T. Fitter and K.~J. Kuchenbecker, ``Designing and assessing expressive
  open-source faces for the baxter robot,'' in \emph{International Conference
  on Social Robotics}.\hskip 1em plus 0.5em minus 0.4em\relax Springer, 2016,
  pp. 340--350.

\bibitem{benson2016modeling}
D.~Benson, M.~M. Khan, T.~Tan, and T.~Hargreaves, ``Modeling and verification
  of facial expression display mechanism for developing a sociable robot
  face,'' in \emph{2016 International Conference on Advanced Robotics and
  Mechatronics (ICARM)}.\hskip 1em plus 0.5em minus 0.4em\relax IEEE, 2016, pp.
  76--81.

\bibitem{8520655}
D.~{Szafir}, B.~{Mutlu}, and T.~{Fong}, ``Communicating directionality in
  flying robots,'' in \emph{2015 10th ACM/IEEE International Conference on
  Human-Robot Interaction (HRI)}, 2015, pp. 19--26.

\bibitem{szafir2015communicating}
D.~Szafir, B.~Mutlu, and T.~Fong, ``Communicating directionality in flying
  robots,'' in \emph{2015 10th ACM/IEEE International Conference on Human-Robot
  Interaction (HRI)}.\hskip 1em plus 0.5em minus 0.4em\relax IEEE, 2015, pp.
  19--26.

\bibitem{bejerano2018methods}
G.~Bejerano, G.~LeMasurier, and H.~A. Yanco, ``Methods for providing
  indications of robot intent in collaborative human-robot tasks,'' in
  \emph{Companion of the 2018 ACM/IEEE International Conference on Human-Robot
  Interaction}, 2018, pp. 65--66.

\bibitem{shrestha2016intent}
M.~C. Shrestha, A.~Kobayashi, T.~Onishi, E.~Uno, H.~Yanagawa, Y.~Yokoyama,
  M.~Kamezaki, A.~Schmitz, and S.~Sugano, ``Intent communication in navigation
  through the use of light and screen indicators,'' in \emph{2016 11th ACM/IEEE
  International Conference on Human-Robot Interaction (HRI)}.\hskip 1em plus
  0.5em minus 0.4em\relax IEEE, 2016, pp. 523--524.

\bibitem{bazilinskyy2019survey}
P.~Bazilinskyy, D.~Dodou, and J.~De~Winter, ``Survey on ehmi concepts: The
  effect of text, color, and perspective,'' \emph{Transportation research part
  F: traffic psychology and behaviour}, vol.~67, pp. 175--194, 2019.

\bibitem{eisma2020external}
Y.~B. Eisma, S.~van Bergen, S.~Ter~Brake, M.~Hensen, W.~J. Tempelaar, and J.~C.
  De~Winter, ``External human--machine interfaces: The effect of display
  location on crossing intentions and eye movements,'' \emph{Information},
  vol.~11, no.~1, p.~13, 2020.

\bibitem{dey2020color}
D.~Dey, A.~Habibovic, B.~Pfleging, M.~Martens, and J.~Terken, ``Color and
  animation preferences for a light band ehmi in interactions between automated
  vehicles and pedestrians,'' in \emph{Proceedings of the 2020 CHI Conference
  on Human Factors in Computing Systems}, 2020, pp. 1--13.

\bibitem{wickens1998introduction}
C.~D. Wickens, S.~E. Gordon, Y.~Liu, and J.~Lee, ``An introduction to human
  factors engineering,'' 1998.

\bibitem{fitzpatrick2006another}
K.~Fitzpatrick, M.~A. Brewer, and S.~Turner, ``Another look at pedestrian
  walking speed,'' \emph{Transportation research record}, vol. 1982, no.~1, pp.
  21--29, 2006.

\end{thebibliography}
\end{document}